\documentclass[preprint,12pt]{elsarticle}

\usepackage{graphicx}
\usepackage[utf8]{inputenc} % allow utf-8 input
\usepackage[T1]{fontenc}    % use 8-bit T1 fonts
\usepackage{hyperref}       % hyperlinks
\usepackage{url}            % simple URL typesetting
\usepackage{booktabs}       % professional-quality tables
\usepackage{amsfonts}       % blackboard math symbols
\usepackage{nicefrac}       % compact symbols for 1/2, etc.
\usepackage{microtype}      % microtypography
\usepackage{microtype}
\usepackage{graphicx}
\usepackage{subfigure}
\usepackage{booktabs} % for professional tables
\usepackage{microtype}
\usepackage{graphicx}
\usepackage{booktabs} 
\usepackage{hyperref}       % hyperlinks
\usepackage{url}            % simple URL typesetting
\usepackage{booktabs}       % professional-quality tables
\usepackage{amsfonts}       % blackboard math symbols
\usepackage{nicefrac}       % compact symbols for 1/2, etc.
\usepackage{microtype}   
\usepackage{diagbox}% microtypography

\usepackage{amsmath}
\newtheorem{theorem}{Theorem}

\newtheorem{definition}{Definition}

\allowdisplaybreaks[4]
\usepackage{amssymb}
\usepackage{subfigure}
\usepackage{graphics}
\usepackage{multirow}
\usepackage{float}
\usepackage{bm}
\usepackage{mathrsfs}
\usepackage{color,soul}
\usepackage{enumerate}
\usepackage{todonotes}
\usepackage{latexsym}
\usepackage{wrapfig} 
\usepackage[normalem]{ulem}
\usepackage{dsfont}
\sethlcolor{yellow}% for professional tables
\usepackage{diagbox}
\usepackage{multirow}
\usepackage{stfloats}
\usepackage{lipsum} 
\usepackage{hyperref} 
\usepackage{algorithm}
\usepackage{algorithmic}
\allowdisplaybreaks[4]
\usepackage{amssymb}
\usepackage{lineno}

\journal{European Journal of Operational Research}

\begin{document}

\begin{frontmatter}


\title{Bilevel Learning Model Towards Industrial Scheduling }

\author{Longkang Li$^1$, Hui-Ling Zhen$^2$, Mingxuan Yuan$^2$, Jiawen Lu$^2$, Xialiang Tong$^2$, Jia Zeng$^2$, Jun Wang$^{2,3}$, Dirk Schnieders$^1$}

\address{1. Hong Kong University. 2. Noah's Arc Lab, Huawei. 3. University College London }

\begin{abstract}
Automatic industrial scheduling, aiming at optimizing the sequence of jobs over limited resources, is widely needed in manufacturing industries. However, existing scheduling systems heavily rely on heuristic algorithms, which either generate ineffective solutions or compute inefficiently when job scale increases. Thus, it is of great importance to develop new large-scale algorithms that are not only efficient and effective, but also capable of satisfying complex constraints in practice. In this paper, we propose a Bilevel Deep reinforcement learning Scheduler, \textit{BDS}, in which the higher level is responsible for exploring an initial global sequence, whereas the lower level is aiming at exploitation for partial sequence refinements, and the two levels are connected by a sliding-window sampling mechanism. In the implementation, a Double Deep Q Network (DDQN) is used in the upper level and Graph Pointer Network (GPN) lies within the lower level. After the theoretical guarantee for the convergence of BDS, we evaluate it in an industrial automatic warehouse scenario, with job number up to $5000$ in each production line. It is shown that our proposed BDS significantly outperforms two most used heuristics, three strong deep networks, and another bilevel baseline approach. In particular, compared with the most used greedy-based heuristic algorithm in real world which takes nearly an hour, our BDS can decrease the makespan by 27.5\%, 28.6\% and 22.1\% for 3 largest datasets respectively, with computational time less than 200 seconds.
\end{abstract}


\end{frontmatter}

%\linenumbers

\section{Introduction} 
 
Scheduling is a synthesized process of arranging, controlling and optimizing task and workload in a manufacturing system~\cite{industry_scheduling}. Typically, one of the most important objectives for scheduling problem is to find the best sequence or permutation of jobs which minimizes the makespan in the limited computational time. Makespan describes the time distance from the start to the end when scheduling a sequence of jobs or tasks. The allocation of machinery resources, planning of human resources and arranging of production processes are all decided by the scheduling result. Since there are an enormous number of jobs needs to be scheduled quickly everyday, even a small improvement of the scheduler can bring significant benefits. A good scheduling algorithm should be able to consider conflicted requirements, as well as to achieve the computational efficiency and the solution quality.  

This paper focuses on an industrial case of scheduling problem, which is derived from automatic warehouse and includes multiple production lines, multiple stages and multiple machines. In general, it is an extended problem from job shop scheduling problem (JSSP)~\cite{industry_scheduling} or flow shop scheduling problem (FSSP)~\cite{flow2017review,flow2019review}. JSSP assign each job to certain resources at particular times, and different jobs may have different precedence constraints. Precedence constraints decide the order of stages or operations when processing a job. FSSP is a special case of JSSP, where all the jobs share the same precedence constraints.

For general scheduling problems, there are three major categories of algorithms. The first category is the heuristic algorithm, such as NEH~\cite{neh1990some} or greedy search~\cite{greedy2007simple}, which is mostly used in practical scenarios. However, the heuristic algorithms depend on local information. The lacking of a global view makes them fail to guarantee the quality of the scheduling solution. When the problem scale increases, some heuristics such as NEH can hardly get feasible solutions even after running a long time. The second category contains the programming models, including mathematical programming~\cite{mip2013flow,cp2012constraint} and dynamic programming~\cite{dynamic2012two}. However, the pure programming models which targeting on optimal solutions only work for small scale problems or problems with very specific structures. Due to the high complexity of industrial scheduling problems, practical mathematical programming solutions are often combined with many heuristic policies~\cite{neh_test2019,neh1990some,greedy2018discrete}. Given a practical problem, operation research toolkits such as Google OR-Tools~\cite{ortools} and OptaPlanner~\cite{opta}, which implement hybrid solutions and provide user-friendly interface to many applications, are often considered firstly.
The third category is reinforcement learning (RL), such as Q-learning (QL)~\cite{flow2017review,ql_test2019two} and policy gradient~\cite{bilevel_rl2018convergent,gpn2019combinatorial}. RL is well-suited to learn scheduling solutions because it allows learning from actual
workload and operating conditions without accurate assumptions. However, off-the-shelf RL algorithms are hard to accurately capture the problem's characteristics and structure, resulting in the inadaptability to the complexity and the scale of industrial scheduling problems. 

\begin{wrapfigure}{r}{0.5\textwidth}
	\includegraphics[width = 0.45\textwidth]{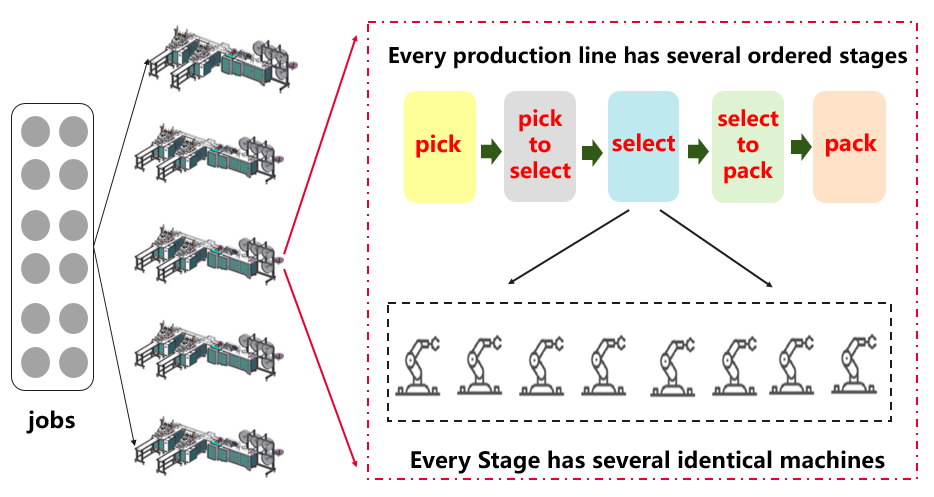} 
	\caption{Industrial Scheduling Flowchart}
	\label{problem_intro} 
\end{wrapfigure}

Considering the above challenges, in this paper, we formulate the industrial scheduling problem as a bilevel constraint Markov Decision Process (MDP), and design a novel bilevel deep reinforcement learning (DRL) method, \textit{Bilevel Deep Scheduler (BDS)} to meet both the efficiency and effectiveness requirements of practical problem. The upper level utilizes a Double Deep Q Network (DDQN) to find an initial sequence of jobs quickly, and the lower level utilizes a graph pointer network (GPN) to refine the sequence under the multiple-machine constraints. Additionally, we also make the following contributions: (1) We give new representations for states and actions, helping decrease the solution space; (2) We adopt a sliding-window sampling mechanism, which connects two levels through partial information sharing to improve the computational efficiency; (3) We give a theoretical guarantee for the convergence, and further conduct extensive experiments in our multi-stage automatic warehouse scheduling scenario with up to 5000 jobs. Comparing with two most used heuristics~\cite{neh_test2019,greedy2018discrete}, other popular deep networks~\cite{pn2016neural,ddqn}, and another bilevel baseline approach, our BDS demonstrates significant advantages on both effectiveness and efficiency; (4) Besides the complex industrial scheduling problem, our BDS can also be utilized for the large scale JSSP or FSSP. And besides DDQN and GPN, our bilevel structure could be set with other deep reinforcement learning models in the specific scenarios. To the best of our knowledge, this is the first work utilizing bilevel DRL to solve large scale industrial scheduling problems. 

The rest of this paper is organized as follows. The properties of industrial scheduling problem and our automatic warehouse scheduling scenario are illustrated in Section~\ref{sec:sec2}. We compare our work with related works in Section~\ref{sec:back}, and describe our bilevel DRL algorithm in Section~\ref{sec:alg}. Theoretical proof of the convergence is given in Section~\ref{sec:theoGuaran}. The experimental results are discussed in Section~\ref{sec:exp} followed by the conclusion in Section~\ref{sec:con}.  

%=========================
\section{Problem Statement}~\label{sec:sec2}

We focus on a frequently appeared industrial scheduling problem, which is a variant of FSSP. The task is to calculate the execution plan of jobs with multiple stages. Each stage corresponds to multiple machines rather than one machine. Given a set of jobs $J$ = $\{1,...,N\}$, a set of stages $S$ = $\{1,...,s\}$, and a set of machines $M$ = $\{1,...,m\}$. The problem has the following properties: 1. Each job has to be processed throughout $s$ stages; 2. All the $N$ jobs share the same precedence constraints; 3. One stage corresponds to multiple machines. A job can be operated by any idle machine in one stage; 4. The machines within a stage are identical, but the machines in different stages might be different;  5. At the same time, one job can only be processed on one machine, and one machine cannot run more than one job.

Specifically, we solve an industrial scheduling problem in the practical automatic warehouses, as demonstrated in Figure~\ref{problem_intro}. There are five independent production lines, A, B, E, M, T, and different line produces different types of products. There are five stages with the precedence constraints in each line: 1 - Pick, 2 - Pick to select, 3 - Select, 4 - Select to pack, 5 - Pack. All the jobs have to be processed sequentially through these five stages. 

Due to the heterogeneity of jobs and productions, we use a clustering algorithm~\cite{cluster2001enhancing} to distribute jobs to production lines and then schedule each production line with our proposed BDS. The inputs of the scheduling problem include the operation times of different jobs at different stages, the precedence constraints and other resource configurations. The output is the job sequence plan (i.e. job sequences on each machine) with the makespan as smaller as possible. It should be emphasized that the job scale evaluated in most academic research works is no larger than $500$, whereas in our practical scenarios, each time there are totally up to $25000$ jobs to be scheduled for all the production lines and the BDS of each production line must deal with up to $5000$ jobs. 

\subsection{Mathematical Model}

The bilevel constrained MDP is a special form of bilevel reinforcement learning~\cite{bi_ac2019bi}, since 1) both their objectives are the summation of the discounted rewards in sequential states, 2) the analytic forms of objectives are unknown and they can only be learned through interactions with the environment in a model-free way. Their difference lies within the constraints which includes both states and policy in our model. Moreover, if we consider $\mathbb{E}_{r_u,r_u,\cdots \sim \pi_u,\pi_u} \left[ \sum_{t=1}^{\propto} \gamma_u^t r_{t,u} \right]$ as leader and $\left[ \sum_{t=1}^{\propto} \gamma_{\ell}^t r_{t,\ell} \right]$ as follower, we can connect Model~(1) to a bilevel game~\cite{game}. As proved in the related works~\cite{interpretable,bilevel_rl2018convergent}, Model~(1) can converge to the Stackelberg equilibrium approximately with certain algorithms.

Now, we are ready to realize Model~(1) for our specific problem. Given a production line with $N$ jobs, $I$ stages and $M_I$ machines in each stage, and we utilize $i$, $j$ and $k$ to label the stage, job and machine, respectively. Our goal is to find the optimal job sequence and job assignment that lead to the minimal makespan $C_{make}$. The inputs also include $O_{i,j,k}$, which denotes the operation time of job $j$ on machine $k$ in stage $i$, the resource configurations and the process constraints. Hereby recall makespan is the total length of the schedule and it is decided by the slowest job. 

Before giving the detailed formulations, let us use an example to explain our design. Supposing job $j$ is the slowest one, thus the tardiness time that it costs decides the total makespan. The tardiness time contains two parts, one is to wait for an available stage, and the other is to wait for an available machine. The former one is caused by waiting for other jobs due to the precedence and non-overlapping constraints. The latter one occurs when waiting for an available machine. Thus, targeting on minimizing the total makespan, we minimize the makespan for stages and that for machines together. Connecting to bilevel model, we have $C_{make} =C_{make,u}+C_{make,\ell}$, where $C_{make,u}$ and $C_{make,\ell}$ denotes the makespans for different stages and machines, respectively. By setting the rewards at time $t$ as $R_{t,u}=1/C_{make,u}$ and $R_{t,\ell}=1/C_{make,\ell}$, we can utilize Eq~(1) to formulate the proposed industrial scheduling problem.

We proceed by having the following definitions: 

\begin{itemize}
\item $s_{i,j}^{(u)}$: $0-1$ state of job $j$ at stage $i$. $s^{(u)}_{i,j}=1$ denotes the job $j$ lies within stage $i$, otherwise $s_{i,j}^{(u)}=0$. 
\item $s_{i,j,k}^{(\ell)}$: $0-1$ state of job $j$ at stage $i$ on machine $k$. $s^{(\ell)}_{i,j,k}=1$ denotes the job $j$ occupies machine $k$ in stage $i$, otherwise $s^{(\ell)}_{i,j,k}=0$.  
\item $v_{j,k}$: starting time of the job $j$ on machine $k$ 
\item $u_{i,j}$: starting time of the job $j$ at stage $i$.
\item $\xi_{i}$: a large positive number to control the intervals between the different jobs on the machines in $i-$th stage. 
\end{itemize}

We then have the specific form of $\Omega_{u}$ for the upper level constraints as 
\begin{align}
& \sum_{i=1}^I s^{(u)}_{i,j}=1, \ \ \forall \ j, \ \ \ \sum_{j=1}^N s^{(u)}_{i,j}\leq 1, \ \ \forall \ i, \label{cons:nonoverlap} \\
& u_{i,j+1} \geq u_{j,i} + \sum_{k}^{M_i} O_{i,j,k}, \ \ \forall \ i,j \leq m, \label{cons:pre_stage} 
\end{align}
while the lower level constraints $\Omega_{\ell}$ are
\begin{align}
&  \sum_{k=1}^{M_i} s^{(\ell)}_{i,j,k} =1, \ \  \forall \ i, j, \ \ \  \sum_i^{I} \sum_j^{N} (s_{i, j, k}) <= 1,\ \ \forall \ k, \label{cons:nonoverlap_3},\\
&  v_{j,k} \geq v_{j,t} + \sum_{t=1}^{k} s^{(\ell)}_{i,j,k} O_{i,j,k}, \ \ \forall k>1, \ t<k \label{cons:pre_machine},\\
&  \left| v_{j,k} - u_{j,i} \right| \leq \xi_i(1- s^{(\ell)}_{i,j,k}), \ \ \forall i, j,k \label{cons:pre_machine_2}, \\
& \frac{1}{R_{i,\ell}}  \geq \sum_{j=1}^N u_{i,j} + \sum_{j=1}^N \sum_{k=1}^{M_i} s^{(\ell)}_{i,j,k} O_{i,j,k}, \ \ \forall i,j. \label{cons:wait_jobs}
\end{align}

Constraint~(\ref{cons:nonoverlap}) ensures the non-overlapping constraint in every stage: (i) each job can only occupy one stage once, (ii) each job can only be assigned to one stage (i.e., the right side is $1$) once or in the pending list (i.e., the right side is $0$). Constraint~(\ref{cons:nonoverlap_3}) ensures the non-overlapping among machines: (i) one machine can only be assigned one job once; (ii) each job can only be assigned to one machine. Constraint~(\ref{cons:pre_stage}) controls the flow and ensures that job $j+1$ cannot be started until the completion of job $j$. Constraint~(\ref{cons:pre_machine}) ensures that job $j$ at stage $i$ cannot begin before the completion of other jobs on the same machine.  Constraint~(\ref{cons:pre_machine_2}) maintains the relationship between each job and each machine in each stage. Constraint~(\ref{cons:wait_jobs}) restricts the tardiness of every job. All variables must be positive numbers.

In the above model, the 0-1 integer variable $s^{(u)}_{i,j}$ and $s^{(\ell)}_{i,j,k}$ are decision variables. 
Different from previous definitions~\cite{industry_scheduling,mip2013flow,cp2012constraint}, there is a sequential relationship between $s^{(u)}_{i,j}$ and $s^{(\ell)}_{i,j,k}$, that is to say, $\forall i$ and $j$, $s^{(\ell)}_{i,j,k}=1$ must be under the assumption of $s^{(u)}_{i,j}=1$. 
The model also shows the relationship between the proposed industrial scheduling problem and job shop or flow shop scheduling problem, i.e., the industrial one can degenerate to either the job shop or flow shop scheduling there only exists one level. 
It is worth noting that our constraints limit the states directly, instead of restricting actions $a_{u}$ and $a_{\ell}$. Thus, in practice, the actions are chosen greedily, as long as the constraints of states are satisfied.

%====================
\section{Related Works and Analysis}~\label{sec:back}

There are three kinds of algorithms for scheduling system. The first are heuristic algorithms, which include iterated greedy algorithm~\cite{greedy_test2019},  NEH~\cite{neh_test2019} and Tabu search~\cite{tabu2016effective}. They are mostly used in practical scenarios. Those methods are comprehensible and explainable in deployments, showing good performance in small scale datasets. However, the heuristics are either failed to meet the solution quality requirements or limited by the sharply decreased computational efficiency, with the job scale and problem complexity increasing. Although some greedy-based methods are computationally acceptable, they usually result in relatively low solution quality. Thus the solution quality and robustness of heuristic algorithms cannot be guaranteed.

\begin{wrapfigure}{r}{0.5\textwidth}
	\includegraphics[width = 0.45\textwidth]{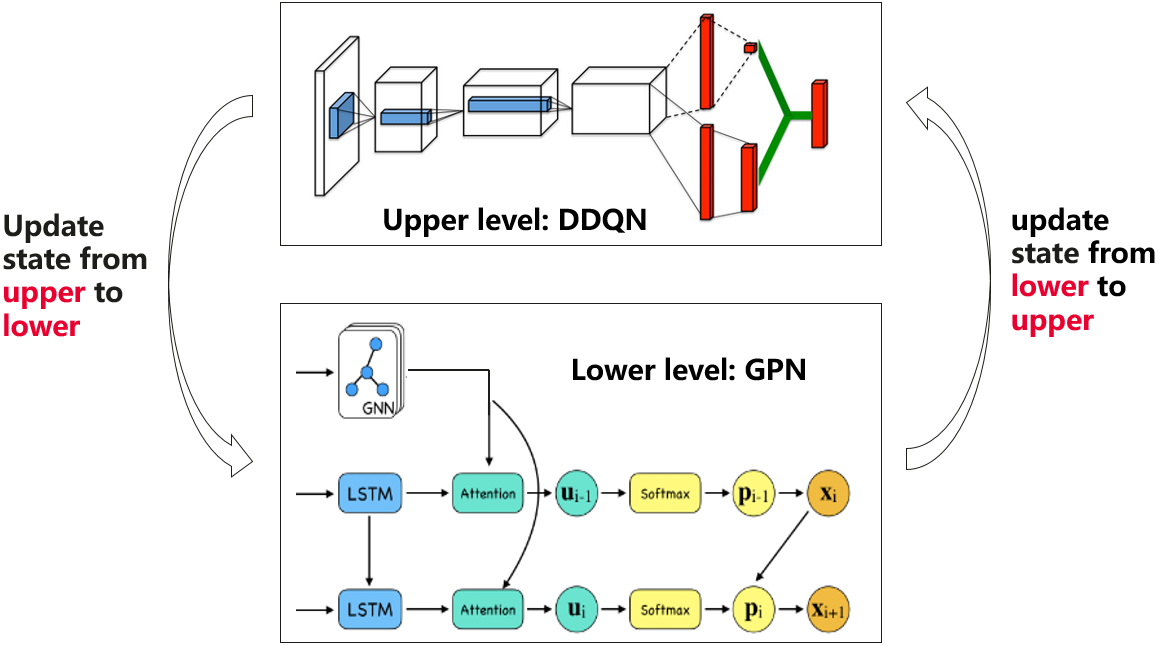} 
	\caption{The illustration of our proposed BDS} 
	\label{fig2}  
\end{wrapfigure}

The second are programming algorithms, including mathematical programming~\cite{mip2013flow,cp2012constraint} and dynamic programming~\cite{dynamic2012two}. Different from heuristics, it is possible to get the analytical and accurate solutions from mathematical models with strong theoretical guarantee. However, pure programming methods only work for small scale cases due to the NP-Hardness of scheduling problems. Some toolkits, which combine the heuristics and programming methods have also been utilized, such as OR-Tools~\cite{ortools} and OptaPlanner~\cite{opta}. These toolkits have implemented user-friendly interfaces for many general logistics applications. However, several tests have shown that they cannot perform well on specific complex problems (more details are given in Appendix I). It can be observed that the computational time of OR-Tools increases rapidly with job number increasing. Even though there are only 30 jobs, the computational time takes over 800 seconds.

The third category are the learning-aided methods, which have been increasingly attracting people's attention in solving scheduling problem recently. Q-learning (QL) is the first RL algorithm to be applied to scheduling problems, and it depends on a guaranteed MDP for finding the optimal policy~\cite{ql_first2000}. And the evaluations of the QL method on our scheduling problem in Section~\ref{sec:sec2} are also given in Tables 1. The discount factor $\gamma$ is set as $0.8$. Similar to OR-Tools, the computational time of QL (more details are given in Appendix I) also increases rapidly. Apparently, when there are over $1000$ jobs in industrial scheduling applications, neither OR-Tools nor Q-learning can meet the requirements of computational efficiency. Besides QL, some extension algorithms combining QL with other search mechanisms, such as tree search, cutting planes and etc.~\cite{flow2019review}, were proposed. People have also been trying to apply RL for many other combinatorial optimization problems. He et al.~\cite{he2014learning} first applied DRL method for some the Traveling Salesman Problem. Bello et al.~\cite{pn2016neural} proposed pointer network (PN) to solve the Vehicle Routing Problem (VRP). Hu et al.~\cite{ali2017solving} then applied a PN for a new 3D bin packing problem in which a number of cuboidshaped items must be put into a bin orthogonally one-by-one. Although great achievements have been made recently, how to apply high efficiency and high quality learning-aided methods in industrial scheduling system is still an open question.

%===========
\section{Our Algorithm}~\label{sec:alg}

In this paper, we propose a \textit{Bilevel Deep Scheduler (BDS)}, for the efficiently schedule a large number of jobs with hierarchy and precedence constrains, we formulate the scheduling problem as a constraint Markov Decision Process (MDP), and design a bilevel deep reinforcement learning scheduler (BDS) to solve the problem. As demonstrated in Figure~\ref{fig2}, BDS contains two levels, the upper level is to explore an initial global sequence of all jobs quickly, and the lower level exploits on the partial sequence for further refinements. The two levels are connected by a sliding-window sampling mechanism, through partial information sharing. The objective of BDS is to schedule all jobs to minimize the makespan. We make use of Double Deep Q Network (DDQN) as the upper level for efficient global sequence initialization and graph pointer network (GPN) as the lower level for high quality solution refining and optimization. It should be emphasized that BDS is a unified framework, in which the upper and lower levels are not restricted to DDQN and GPN. People can freely choose other adaptive models in their scenarios.

\subsection{Bilevel Constraint MDP}

Constrained MDP can be modeled as a five tuple $(\mathcal{S},\mathcal{A}, P, R, \Omega)$, where $\mathcal{S}$ denotes the state space with a limited number of state $s$, $\mathcal{A}$ is the action space consisting of a series of action $a$, $R$ is the reward function incurred when executing action $a$ from state $s$, $\Omega$ is the set of constraint functions from the industrial manufacturing scheduling scenario, and $P$ is transition probability from one state to another, where $P: \mathcal{K} \times \mathcal{S} \rightarrow [0,1]$, with $\mathcal{K}=\{(s,a)|s\in \mathcal{S},a\in \mathcal{A}\}$.

Then, we can model the bilevel constrained MDP via two five tuples $(\mathcal{S}_u,\mathcal{A}_u, P_u, R_u, \Omega_u)$ and $(\mathcal{S}_{\ell},\mathcal{A}_{\ell}, P_{\ell}, R_{\ell}, \Omega_{\ell})$. Here, the subscript $u$ denotes the upper level and $\ell$ is the lower level.
Let $\pi_u(a_{u}|s_{u})$ and $\pi_{\ell}(a_{\ell}|s_{\ell})$ be two policies for the two levels respectively,  and $\gamma$ be the discount, the bilevel constrained MDP is formulated as 

\vspace{-2ex}
\begin{align}
    & \pi_{u}^*, s_{u}^* = \mathrm{argmax}_{\pi_u,s_u} \mathbb{E}_{R_{1,u},R_{2,u},\cdots \sim \pi_u,\pi_{\ell}^*} \left[ \sum_{t=1}^{\propto} \gamma_u^t R_{t,u} \right] \label{eq:upper_level} \\
& s.t., \pi_u, s_u \in \Omega_u, \nonumber  \\
& \ \ \ \ \ \ \pi_{\ell}^*,s_{\ell}^* = \mathrm{argmax}_{\pi_{\ell},s_{\ell}}  \mathbb{E}_{R_{1,\ell},R_{2,\ell},\cdots \sim \pi_{u}^*,\pi_{\ell}} \left[ \sum_{t=1}^{\propto} \gamma_{\ell}^t R_{t,\ell} \right] \label{eq:lower_level}  \\
& \ \ \ \ \ \ \ s.t., \pi_{\ell}, s_{\ell} \in \Omega_{\ell}. \nonumber
\end{align}
Due to space limitations, we put the detailed descriptions in Appendix II.

%===================================================
\subsection{DDQN for Upper Level}

We utilize a DDQN~\cite{ddqn} to solve the upper level model~(\ref{eq:upper_level}), aiming to explore an initial global sequence among different stages. The input is the operation times of each stage of each job on specific machines. The structure of DDQN is demonstrated in Figure~\ref{fig2}.

We design the following RL model: (i) \textit{State} $\mathbf{s}_{u}$ denotes the current job sequence. (ii) \textit{Action} $\mathbf{a}_{u}$ denotes the selected job which will be processed next. (iii) \textit{Reward} $= 1/M$,  where $M$ denotes the makespan of current job sequence. The smaller the makespan, the larger the reward. (iv) \textit{Policy} $\pi_{{\theta},{\theta^{*}}}(\mathbf{a}_{u} | \mathbf{s}_{u})$ is a distribution over all candidate jobs, where ${\theta},\theta^{*}$ represents the weights of DDQN. 
Using the above state and action definitions, the computational complexity to get the job sequence can be decreased a lot from $\mathcal{O}\left(n!\right)$~\cite{n_flow} or $\mathcal{O}\left(2^n\right)$~\cite{ql_first2000}.

Algorithm~3 demonstrates the outer loops where the upper level DDQN is co-trained with the lower level GPN together. DDQN starts with initial state $s_u = empty$ in the first round. Then in the following co-training loops, it inputs the refined sequence from the lower level, and it firstly updates the network according to the input $s_u$, then resets to empty list. The network stops updating when both models in two levels converge. We show the inner training process of DDQN in Algorithm~1. DDQN always starts with initial state $s_u = empty$, then recursively adds a new job into the list until getting a new sequence with size $N$.  

\begin{wrapfigure}{r}{0.5\textwidth}
\includegraphics[width = 0.5\textwidth]{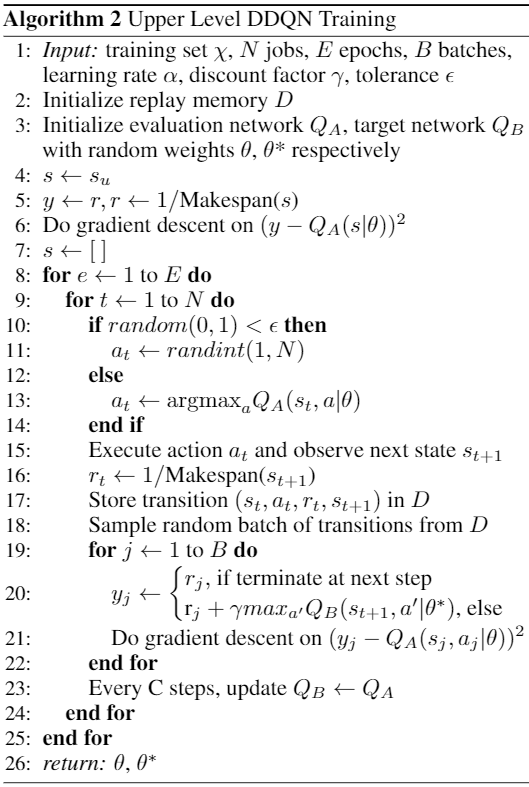}\label{alg1}
\end{wrapfigure} 

The DDQN is co-trained with the lower level GPN together. DDQN starts with initial state $s_u = empty$ in the first round. Then in the following co-training loops, it inputs the refined sequence from the lower level. We show the co-training process of DDQN in Algorithm~1, where DDQN firstly updates the network according to the input $s_u$, then it resets to empty list and recursively adds a new job into the list until getting a new sequence with size $N$. The network stops updating when both models in two levels converge. 

As shown in Algorithm~3 as outer loops, DDQN starts with the initial sequence $s_u = [1,2,..,N]$ in the first loop, while in the other loops it inputs the refined sequence from the lower level. Algorithm~1 shows the training process as an inner loop, where DDQN firstly updates the network according to the input $s_u$, secondly resets to empty list, and then recursively adds a new job into the list until reaching size $N$. The network keeps updating as the learning proceeds. 

During testing, the upper level finishes its calculation when a state with size $N$ is obtained and delivers the result (initial global sequence) to the lower level for further refinements.

%==================================================
\subsection{GPN for Lower Level}

We propose to solve lower level model~(\ref{eq:lower_level}) via a GPN~\cite{gpn2019combinatorial}, aiming at the exploitation for the partial  sequence refinement. Different from DDQN, its input is the operation times of a batch of jobs.

After given the following definitions similar to the upper level, such as  
\textit{State} $\mathbf{s}_{\ell}$, 
\textit{Action} $\mathbf{a}_{\ell}$, 
\textit{Policy} $\pi_{\theta'}(\mathbf{a}_{\ell}| \mathbf{s}_{\ell})$, GPN can be trained via the layer-wise policy gradient method. Algorithm 2 shows the corresponding training procedure, where $\theta'$ refers to the weights. 

During testing, the GPN outputs a refined partial sequence, and bring it back to the initial global sequence for updating. The structure of GPN is outlined in Figure~\ref{fig2}. Clearly, there are two components in GPN:

\begin{wrapfigure}{r}{0.5\textwidth}
\includegraphics[width = 0.5\textwidth]{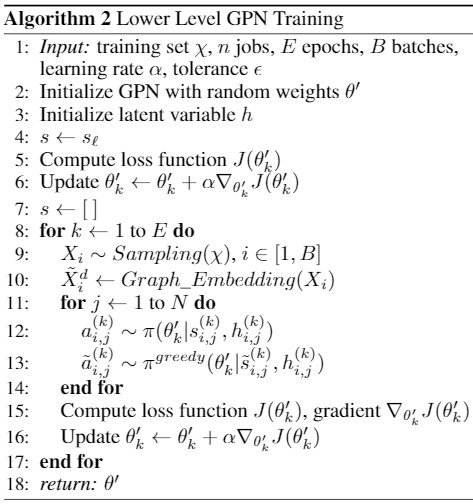}\label{alg2}
\end{wrapfigure} 

(1) \textit{Graph embedding and LSTM encoder}. Let $x_{i,j} \in \mathbb{R}^s$ be the operation time for job $i$ in stage $j$, where $i=\{1,..,s\}, j=\{1,..,n\}, n<N$, $s$ refers to dimension. Using graph embedding, $x_{i,j}$ is embedded into $\tilde{x}_{i,j} \in \mathbb{R}^d$, where $d$ is the hidden dimension, $d > s$. Then the vector $\tilde{x}_{i,j}$ is encoded via LSTM. Meanwhile, the hidden variable $h^{\tilde{x}_{i,j}}$ of the LSTM is passed to both the decoder in the current step and the encoder in the next time step. 

(2) \textit{Attention-based Decoder}. It outputs a pointer vector $u_i$, which is then passed to a softmax layer to generate a distribution over the next candidate job.

%Each action $a_{t,\ell}$ is also provided by exploration and exploitation. 

%===================================================
\subsection{\textbf{Sliding-window Interaction}} 

We utilize a sliding-window sampling mechanism to connect the two levels. Algorithm 3 shows our co-training algorithm. 

% Hereby, $A$ and $S$, respectively, denote action and state sets. 

\begin{wrapfigure}{r}{0.5\textwidth}
\vspace{-1ex}
\includegraphics[width = 0.5\textwidth]{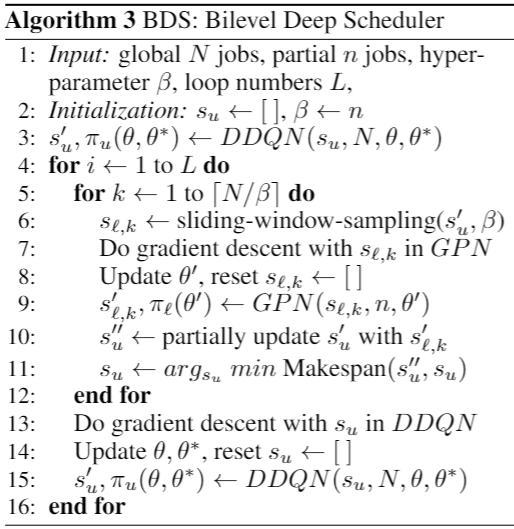}\label{alg3}
\end{wrapfigure} 

There is a recursive loop in our BDS. (i) From the upper to the lower level: DDQN starts with initial state $s_u = empty$ in the first round. Then in the following co-training loops, it inputs the refined sequence $s_u$ from the lower level, updates the network according to the input $s_u$ and runs Algorithm 1 again. DDQN outputs an initial global sequence $s'_u$ with size $N$. After sampling, GPN receives the partial sequence $s_\ell$ with size $n$. $\beta$ is a hyperparameter determining the size of the sequence $s_\ell$, $\beta=n<N$. 
(ii) From the lower back to the upper level: GPN outputs the refined partial sequence $s_\ell'$, and brings it back to the initial global sequence $s'_u$ for updating. The updating mechanism is: if the updated sequence achieves smaller makespan, then update; otherwise, keep the original sequence and continue the next loop. 

To converge to the optimal sequence, Algorithm 3 needs to run a certain number of loops. We use $L$ to represent the number of training loops. The more loops we run, the closer we can get to the optimal solution. The value of $L$ controls the tradeoff between solution quality and computational time.

%=================

\section{Theoretical Guarantee}\label{sec:theoGuaran}

\subsection{Basic Introduction of Game}

We clearly differentiate Stackelberg equilibrium from Nash one, where $P$ and $Q$ correspond with the optimal strategies of players $A$ and $B$:  
%while 

% \begin{figure}[ht]
% \centering
% \subfigure[Stackelberg Equilibrium]{\includegraphics[width=0.49\linewidth]{stack_equ.jpeg}}
% \subfigure[Nash Equilibrium]{\includegraphics[width=0.49\linewidth]{nach_equ.jpeg}}
% %\includegraphics[width=0.49\linewidth]{Figures/nash_equ.jpeg} 
% \caption{Level curves for Stackelberg and Nash equilibriums. Horizontal and vertical coordinates respectively refer to players $A$ and $B$. Here, $P$ and $Q$ respectively represent the global maximum of $F_A$ and $F_B$.  }
% \label{fig:game_compare}
% \end{figure}

\begin{itemize}
\item if the leader $A$ chooses the strategy $a$ firstly, then the optimal solutions of $F_B$ can be expressed as $f(a)$ which goes across its global optimal strategy $Q$, and Stackelberg equilibrium $(a_S,b_S)$ denotes the tangent of $F_A$ and $f(a)$. Obviously, the follower will choose the strategy which is most favorable to leader. 

\item if players $A$ and $B$ do not share information before decision, $(a_N,b_N)$ is a Nash equilibrium since $F_A$ has a horizontal tangent at this point, while $F_B$ has a vertical tangent. It denotes that one cannot increase his payoff by single-mindedly changing his own strategy, as long as the other sticks to the Nash equilibrium.  
\end{itemize}

Considering the optimization process, the leader $\mathcal{A}$ (the upper level objective) adopts the strategy $a$, the follower $\mathcal{B}$ (the lower level objective) requires  to maximize $F_B(a,b)$ and chooses a best reply $b^* = f(a)$, the goal of leader is now to maximize $F_A(a,f(a))$.  
assuming that player $A$ servers as the leader and announces his strategy in advance, then player $B$ makes his choice accordingly. In Pareto optimality, one cannot increase its own payoff strictly without decreasing the payoff of the other.

It is a zero-sum game when $F_A(a,b) + F_B(a,b) =0$.
Here, $A$ and $B$ denote the feasible region of $F_A$ and $F_B$, respectively,
\begin{definition}\label{def:pareto_opt}
A pair of strategies $(a^*,b^*)$, or called feasible solutions of $F_A$ and $F_B$, are \textbf{Pareto optimal} if for every pair $(a,b) \in A \times B$ such that 
\[ F_A(a,b) < F_A(a^*,b^*), \ \mathrm{and} \ F_B(a,b) \leq F_B(a^*,b^*) \]
or 
\[ F_A(a,b) \leq F_A(a^*,b^*), \ \mathrm{and} \ F_B(a,b) < F_B(a^*,b^*). \]
\end{definition}

\subsection{Proof and Discussions}

For a practical scheduling algorithm, an essential problem is to guarantee the convergence, which is related to the interpretability and robustness~\cite{interpretable}. Aiming at this, we revisit BDS from a game perspective: in this game, DDQN works as a leader and it is to explore the initial solution, while GPN is a follower and its goal is for optimization and refinement. That is to say, DDQN and GPN construct a bilevel game, and the optimal solution of BDS is Stackelberg equilibrium. 

Apparently, the sliding-window sampling plays the role of partial information exchange between two players. Considering above, we can give the following theorem. Besides the contents below, more discussions are given in Appendix III.

\begin{theorem}\label{theo:stack} 
BDS can finally converge to point $E$, and $E= \mathcal{S}_{E}$ (Stackelberg equilibrium) when the weights of BDS meet given conditions.
\end{theorem}

$Proof.$ To prove Theorem~\ref{theo:stack}, we revisit the two levels of BDS as two players. To simplify the label, let $u_1$ denote the upper level DDQN and $u_2$ denotes the lower level GPN. Next, we will prove Theorem~\ref{theo:stack} according to the definition of Stakelberg equilibrium. That is to say, our goal is to prove $\frac{\partial \mathcal{H}_{stack,u_1}}{\partial u_1} = \frac{\partial \mathcal{H}_{stack,u_2}}{\partial u_2}$, where $\mathcal{H}_{stack,u_1}$ and $\mathcal{H}_{stack,u_2}$ are respective Hamiltonians for $u_1$ and $u_2$~\cite{game}. 

We firstly formulate the two Hamiltonians $\mathcal{H}_{stack,u_1}$ and $\mathcal{H}_{stack,u_2}$. Before this, we give the cost functions for $u_1$ and $u_2$, respectively, as follows: 

\vspace{-2ex}
\begin{align*}
& J_1(x,u_1,u_2) = \frac{1}{2} \int_0^{\infty} r_{1}(x,u_1,u_2) r_{1}(x,u_1,u_2)^T \mathrm{d}t,  \\
& J_2(x,u_1,u_2) = \frac{1}{2} \int_0^{\infty} r_{2}(x,u_1,u_2) r_{2}(x,u_1,u_2)^T \mathrm{d}t, 
\end{align*}
where $r_1(x,u_1,u_2) \equiv A_1 x+ B_{11} u_1 - B_{12} u_2$ and $r_2(x,u_1,u_2)  \equiv A_2 x- B_{21} u_1 + B_{22} u_2$. Hereby, $A_i$ and $B_{ij}$ are the learned constants matrices with corresponding dimensions, and the vector $x$ denotes the hyperparameters in bilevel deep networks. 
Since $AA^T$ is a symmetric matrix, we have 
\begin{align*}
&  r_{1}(x,u_1,u_2) r_{1}(x,u_1,u_2)^T=  x^T Q_1 x + u_1^T R_{11} u_1- u_2^T R_{12} u_2, \\ 
& r_{2}(x,u_1,u_2) r_{2}(x,u_1,u_2)^T =  x^T Q_2 x - u_1^T R_{21} u_1 + u_2^T R_{22} u_2,
\end{align*}
where $Q_j$ and $R_{jk}$ are both positive definite and symmetric. Thus, the optimal cost functions for $u_1$ and $u_2$ can be constructed in the following forms: 
\begin{align}\label{eq:optimal_cost}
\begin{split}
& J_1^*(x,u_1,u_2) = \mathrm{min}_{u_1} \frac{1}{2} \int_t^{\infty} r_1(x^*,u_1^*,u_2^*) \mathrm{d} t,\\
& J_2^*(x,J_1^*,u_2) =  \mathrm{min}_{u_2} \frac{1}{2} \int_t^{\infty} r_2(x,J_1^*,u_2) \mathrm{d} t. 
\end{split}
\end{align}
Notice that the optimal cost function of $u_2$, i.e., $J_2^*(x,J_1^*,u_2)$, is affected by that of $u_1$, which is also the main differences between Nash and Stackelberg games~\cite{game}.

Secondly, we compute the two gradients. According to Equation~(\ref{eq:optimal_cost}), we can give the gradient of \textit{ leader's Hamiltonian} as  

\[ \mathcal{H}_{stack,u_1} =  r_1(x,u_1,u_2) +   \left( \nabla J_1 \right)^T (A_1 x+ B_{11} u_1- B_{12} u_2).  \]
Through $\frac{\partial \mathcal{H}_{stack,u_1}}{\partial u_1}=0$, the optimal controller for leader can be given by $u_1^* = - \frac{1}{2} \left( R_{11} \right)^{-1} B_{11}^{T} \nabla J_1$, as well as $\frac{\partial \mathcal{H}_{stack,u_1}}{\partial x} = (Q_1 + Q_1^T) x + A^T \nabla J_1$. Therefore, the optimal cost of $u_2$ can be rewritten as 
\[ J_2^*(x,J_1^*,u_2)  =  \mathrm{min}_{u_2}  \frac{1}{2} \int_t^{\infty} \bigg[ x^T Q_2 x + \left( u_1^* \right)^T R_{21} u_1^*
+ u_2^T R_{22} u_2 + \lambda  \frac{\partial \mathcal{H}_{stack,u_1}}{\partial x} \bigg] \mathrm{d} t, \] 
where $r_2(x,J_1^*,u_2)=r_2(x,u_1,u_2)\bigg|_{u_1=J_1^*}$.
It means that the Hamiltonian of follower $u_2$ can be rewritten as 
\[ \mathcal{H}_{stack,u_2} =  r_2(x,u_1,u_2) +  \left( \nabla J_2 \right)^T (A_2 x - B_{21} u_1 + B_{22} u_2).  \]
Then we can compute the gradient of $\mathcal{H}_{stack,u_2}$ as 
\[ \mathcal{H}_{stack,u_2} =  \gamma \left[ (Q_1 + Q_1^T) x + A^T \nabla J_1 \right]  + r_2(x,J_1^*,u_2) + \left(\nabla J_2 \right)^T (A x + B_1 J_1^* + B_2 u_2). \]

Finally, we can find that when $u_2^* =  - \frac{1}{2} \left( R_{22} \right)^{-1}  B_{22}^{T} \nabla J_2$, we have 
\[ r_2(x,J_1^*,u_2) =  x^T Q_2 x + \left( u_1^* \right)^T R_{21} u_1^* + u_2^T R_{22} u_2 + \gamma \frac{\partial \mathcal{H}_{stack,u_1}}{\partial x}. \]
Assuming that $B_{11} u_1 + B_{22} u_2 =0$, we can obtain that 
\[ \langle u_1 R_{11}^T R_{22} u_2, \left( \nabla_u \mathcal{H}_{stack} \right)^T \rangle = 0, \ \ i.e., \ \   \frac{\partial \mathcal{H}_{stack,u_1}}{\partial u_1} =\frac{\partial \mathcal{H}_{stack,u_2}}{\partial u_2} =0. \]
That is to say, BDS can converge after several loops, and the leader and follower (i.e., DDQN and GPN) can converge to the Stackelberg equilibrium under given conditions. 
$\hfill\blacksquare$

%==============
\section{Experimental Results}~\label{sec:exp} 

\vspace{-5ex} 
\subsection{Setting up} 

Firstly, we evaluate our BDS on a set of benchmarks~\cite{opensource} consisting of 8 sub-datasets, as shown in Table~\ref{tab:benchmark}. The job numbers range from 100 to 800. The proportion of training, validation and testing sets is 8:1:1. There are 20 stages and each stage contains 10 identical machines in this benchmark. Then we evaluate our BDS for each production line under the automatic assembling warehouse system as demonstrated in Section~\ref{sec:sec2}. Each production line contains 5 stages and each stage consists of 10 machines. We test two datasets by setting the operation times of jobs follow uniform distribution and $\chi^2$ distribution respectively. Each dataset contains $6$ sub-datasets with scales of $100$, $250$, $500$, $1000$, $2000$ and $5000$ jobs. The proportion of training, validation and testing sets are also set as 8:1:1.

We compare our BDS with six baseline algorithms, including two most used heuristic algorithms and four deep learning based algorithms. Heuristic$~1$ is a greedy-based~\cite{greedy_test2019} algorithm which is claimed to be the most used algorithm in real world for finding the minimal makespan in FSSP. Heuristic$~2$ is based on NEH~\cite{neh_test2019} and it is claimed to outperform the other heuristic algorithms. DDQN~\cite{ddqn}, PN~\cite{pn2016neural}, GPN~\cite{gpn2019combinatorial} are recently proposed deep learning models for combinatorial optimizations. Limited by the computation resources, we train PN and GPN with 50 jobs and test them on larger datasets. Besides the above five algorithms, we also compare BDS with another bilevel model, whose upper level is DDQN and lower level is PN. We study two core criteria, makespan (scheduling effectiveness) and computation time (efficiency). More discussions on the impact of hyperparameters $\beta$ is put in the Appendix IV.

The experiments are conducted on a server with Centos $7$, $x64$, Intel $E7-4820$, $64$ CPU Cores and one $Tesla$ $V100$-SXM2 GPU. The number of outer loops $L$ is set as 2. Sliding window size $\beta$ is set as 100, 50 and 100 in the benchmark and two automatic warehouse datasets, respectively. There are at least $500$ epochs in every training procedure, batch size is set as 200 on both levels. We choose $Adam$ as the optimizer. 

\subsection{Basic Comparison} 

It can be observed from the Table~\ref{tab:table1} that the computational time of OR-Tools increases rapidly with job number increasing. Even though there are only 30 jobs, the computational time takes over 800 seconds. Similar to OR-Tools, Q-learning (QL) also increases rapidly in the computational time. Apparently, when there are over $1000$ jobs in industrial scheduling applications, neither OR-Tools nor Q-learning can meet the requirements of computational efficiency.

\begin{table}[!h]
 	\label{tab:table1}
 	\centering
 	\caption{Computational Evaluation on Q-learning and OR-Tools}
	\resizebox{8.5 cm}{!}{
	\begin{tabular}{c|ccc}
		\toprule
		Jobs & Machine(s) per Stage & QL Time & OR-Tools Time\\
		\midrule
		15 & 1 & 16.61s  &  42.1s   \\
		15 & 10 & 145.9s & 65.3s  \\
		20 & 1 & 24.72s & 97.9s \\
		20 & 10 & 263.2s & 139.6s \\
		25 & 1 & 112.8s &  435.5s \\
		25 &  10 & 666.8s & 658.4s \\
		30 & 1 & 1557.2s & 810.8s \\
		30 & 10 & 2853.5s& 1240.5s \\
		\bottomrule
	\end{tabular}}
\end{table}

\subsection{Evaluation on BDS}

\subsubsection{Solution Quality (Effectiveness)}

The makespans of different algorithms are shown in Table~\ref{tab:benchmark}, Table~\ref{tab:solution_data_1} and Table~\ref{tab:solution_data_2}. From the results it can be observed that Heuristic 1 performs the worst, while BDS demonstrates its advantage and obtains the smallest makespan. With job scale increasing, BDS’s advantage increases prominently. We can observe that compared with Heuristic 1, BDS can decrease the makespan by 27.5\%, 28.6\% and 22.1\% respectively for the three largest datasets (800 jobs in benchmark and 5000 jobs in the two warehouse datasets). DDQN-PN can also get makespans close to BDS, which verifies that the bilevel structure helps to get high quality solution.

\subsubsection{Computational Time (Efficiency)}

From the results on the two warehouse datasets in Table~\ref{tab:solution_data_1} and Table~\ref{tab:solution_data_2}, we can observe that Heuristic 1 and DDQN cannot generate good enough results even after running a long time. When $\mathrm{job\ number} \geq 1000$, Heuristic 2 costs more than $200000$ seconds, which is unacceptable in practical scenarios. When $\mathrm{job\ number} = 5000$, PN also runs more than $17,000$ seconds, which is too long to be practically used. Two bilevel solutions, DDQN-PN and BDS, both show great advantage in computation efficiency. Furthermore, BDS runs significantly faster than DDQN-PN. It take less than $200$ seconds, which is only 1/3 running time of DDQN-PN, when $\mathrm{job\ number} = 5000$. With job scale increasing, BDS is with quasi-linear growth in the computational time. Similar results can also be observed on benchmark dataset (Table 1). Considering the computation efficiency and effectiveness, $BDS$ significantly outperforms all the other baseline algorithms, especially for large-scale data. Notice that our test is performed on the datasets with job number $> 50$, which also shows that BDS has better generalization ability than other deep models.

\begin{table}[!t]
	\caption{Performance Comparison on Benchmarks }
	\label{tab:benchmark}
	{\small
		\resizebox{14cm}{!}{
	\begin{tabular}{|c|cc|cc|cc|cc|cc|cc|cc|}
		\toprule
	\multirow{2}*{Dataset}  & \multicolumn{2}{|c|}{Heuristic 1} &  \multicolumn{2}{|c|}{Heuristic 2} & \multicolumn{2}{|c|}{DDQN} & \multicolumn{2}{|c|}{PN} & \multicolumn{2}{|c|}{GPN} & \multicolumn{2}{|c|}{DDQN-PN} & \multicolumn{2}{|c|}{BDS}  \\
	~ & Makespan & Time & Makespan & Time & Makespan & Time & Makespan & Time & Makespan & Time & Makespan & Time & Makespan & Time \\
		\midrule 
		L\_100\_20 & 12379 & 3.46s & 7665 & 99.89s  & 10769 & 51.83s   & 7297 & 42.59s & 7236 & 13.32s &{7251}& 29.46s  & \textbf{7211} & 16.38s  \\
		L\_200\_20  & 22526 & 10.11s & 18918 &1255.11s  & 20979& 66.19s   & 16988&50.56s  & 16824 & 24.07s &{15068}& 42.19s  & \textbf{14793} & 20.19s  \\
		L\_300\_20  & 32287 & 20.11s & 28423 & 6505.81s & 31831 &82.04s    & 26476&71.29s & 26358& 40.24s &{22870}& 50.71s& \textbf{21945} & 47.45s \\
		L\_400\_20  & 42256 & 34.85s & 38310& 18033.72s & 39788& 94.38s    & 36299&124.55s& 36290&60.38s&{30197}&59.33s& \textbf{29485} & 61.03s \\
		L\_500\_20  & 52823 & 54.66s &48794& 48259.65s & 51163&103.96s   & 46821 & 136.99s&46838&91.86s &{37423}&81.29s&\textbf{37616}& 69.35s \\
		L\_600\_20  & 62665 & 73.56s & 58601& 89759.51s & 60128&107.43s& 56665& 126.59s & 56635 & 229.73s& {44688}&84.05s&\textbf{44302}&72.89s\\
		L\_700\_20  & 72289  & 104.04s & 68219   & 154344.34s & 69077&109.11s& 66333&308.65s& 66300 & 162.87s & {51829}& 98.20s&\textbf{51783}&80.98s\\
		L\_800\_20  & 82495  & 131.122s & 78379 & 262408.74s& 79921&128.17s & 76454&355.07s&76359&209.82s &{60687}&106.92s&\textbf{59785}&87.23s\\
		\bottomrule
	\end{tabular}}}
\end{table}

\begin{table}[!t]
	\caption{Performance Comparison on Dataset I }
	\label{tab:solution_data_1}
	{\small
		\resizebox{14cm}{!}{
		\centering
	\begin{tabular}{|c|cc|cc|cc|cc|cc|cc|cc|}
		\toprule
	\multirow{2}*{Job No.}  & \multicolumn{2}{|c|}{Heuristic 1} &  \multicolumn{2}{|c|}{Heuristic 2} & \multicolumn{2}{|c|}{DDQN} & \multicolumn{2}{|c|}{PN} & \multicolumn{2}{|c|}{GPN} & \multicolumn{2}{|c|}{DDQN-PN} & \multicolumn{2}{|c|}{BDS}  \\
	~ & Makespan & Time & Makespan & Time & Makespan & Time & Makespan & Time & Makespan & Time & Makespan & Time & Makespan & Time \\
		\midrule 
		100  & 27.5 & 0.91s & 21.1 & 26.0s & 25.04&44.87s    & 20.88&6.6s & 21.05& 5.08s &{18.78}&35.25s& \textbf{18.16} & 22.23s \\
		250  & 64.56 & 4.55s & 57.88& 627s & 62.77& 64.65s    & 57.79&33.61s&57.65&11.19s&{46.57}&43.37s& \textbf{45.21} & 31.02s \\
		500  & 127.93 & 14.07s &120.6& 17264s & 126.4&87.08s   & 119.72&155.03s&119.74&28.35s &{92.68}&64.16s&\textbf{91.57}& 43.22s \\
		1000  & 252.04 & 52.07s & 244.15& 232997s & 250.9&123.1s& 244.01&634.67s&244.11&92.28s& {181.31}&137.06s&\textbf{179.5}&50.52s\\
		2000  & 499.84  & 224.64s & /   & >500000s & 498.5&154.7s& 491.41&4021.87s&491.38&331.56s&{360.13}&215.01s&\textbf{358.39}&84.38s\\
		5000  & 1255.08  & 3950.38s & / & >1000000s& 1253.6&362.7s & 1245.9&17123.2s&1242.6&  1573.32s      & {910.22} & 508.02s & \textbf{895.96}& 168.11s\\
		% 		30 & 9.43 & 6.35  & 6.97 &  6.14  & \textbf{5.16}   \\
		\bottomrule
	\end{tabular}}}
\end{table}

\begin{table}[!t]
	\caption{Performance Comparison on Dataset II }
	\label{tab:solution_data_2}
	{\small
		\resizebox{14cm}{!}{
		\centering
	\begin{tabular}{|c|cc|cc|cc|cc|cc|cc|cc|}
		\toprule
	\multirow{2}*{Job No.}  & \multicolumn{2}{|c|}{Heuristic 1} &  \multicolumn{2}{|c|}{Heuristic 2} & \multicolumn{2}{|c|}{DDQN} & \multicolumn{2}{|c|}{PN} & \multicolumn{2}{|c|}{GPN} & \multicolumn{2}{|c|}{DDQN-PN} & \multicolumn{2}{|c|}{BDS}  \\
	~ & Makespan & Time & Makespan & Time & Makespan & Time & Makespan & Time & Makespan & Time & Makespan & Time & Makespan & Time \\
		\midrule 
		100  & 65.5 & 0.97s & 39.31 & 26.0s & 57.1          & 55.48s  & 38.69   &6.47s  & 38.55 & 7.74s &{37.66}&42.77s& \textbf{37.45} & 27.89s \\
		250  & 136.34 & 4.1s & 104.62& 627s & 131.55        & 89.07s    & 106.84&34.08s & 106.14&12.39s &{99.21}&56.5s& \textbf{97.63} & 37.42s \\
		500  & 265.84 & 15.94s &227.98& 17264s & 255.15     & 111.77s   & 233.87&162.61s& 228.42&29.64s &{199.01}&93.9s&\textbf{186.32}& 42.81s \\
		1000  & 504.61 & 61.55s & 461.32& 232997s & 497.87  & 160.91s & 476.48 & 794.65s& 462.16s&91.87s&{410.29}&158.93s&\textbf{399.28}&62.52s\\
		2000  & 1013.14  & 235.6s & /  & >500000s & 1006.91 & 188.66s & 978.76 &4091.7s& 972.54 &355.34s&{800.39}&275.06s&\textbf{790.95}&96.43s\\
		5000  & 2478.99  & 3204.2s & / & >1000000s& 2371.01 & 525.7s & 2243.53&39678.53s&2230.12 &1858.9s           &{1954.77}&675.37s&\textbf{1930.01}&171.79s\\
		\bottomrule
	\end{tabular}}}
\end{table}

%=======================
\subsection{Hyperparameter Analysis}

It is clear that the sliding-window sampling matters for total design in BDS. In this section, we aim to explore how the different sliding-window sizes affect the scheduler's performance. We first evaluate the solution quality (makespans) and the computational time of different methods, as shown in Table~\ref{tab:hyperparameter_1} and Table~\ref{tab:hyperparameter_2} . Here, we use $\beta$ to denote the size of sliding-window. The results show that when job scale is larger than $100$, sliding-window sampling can lead to better solution quality than full information sharing, and when $\mathrm{job\ number}\geq 1000$, the sliding-window sampling can bring at least $15\%$ decrease of makespan, compared to other heuristic algorithms. And we find that there is a $\beta$ value, where the best makespan can be achieved. In Table~\ref{tab:hyperparameter_1}, the best $\beta$ is somewhere near 50, while in the Table~\ref{tab:hyperparameter_2}, the best $\beta$ is somewhere near 100. Then we evaluate the test efficiency (computational efficiency of BDS). We can observe that a smaller $\beta$ value may obtain smaller overall computational time. 

\begin{table}[!t]
	\caption{Evaluation on Sliding-window Sampling (Dataset I) } 
	\label{tab:hyperparameter_1}\centering
		{\small
		\resizebox{10.5cm}{!}{
		
	\begin{tabular}{|c|cc|cc|cc|cc|}
		\toprule
		
		\multirow{2}*{Job No.}  & \multicolumn{2}{|c|}{$\beta=25$} &  \multicolumn{2}{|c|}{$\beta=50$} & \multicolumn{2}{|c|}{$\beta=100$} & \multicolumn{2}{|c|}{$\beta=250$}\\
	~ & Makespan & Time & Makespan & Time & Makespan & Time & Makespan & Time \\
		
		\midrule
		100& 23.66 & 21.58s& \textbf{18.16} & 22.23s & 21.05 &27.92s  & / & /\\
		250& 57.3  &34.96s  & \textbf{45.21}& 31.02s & 51.35 & 38.28s & 57.65 & 29.37s\\
		500& 118.5 &37.3s  & \textbf{91.57} & 43.22s & 104.57 & 41.52s & 114.41 & 46.33s\\
		1000&238.23&55.35s &\textbf{179.5}& 50.52s &206.54 & 61.43s  & 229.04 & 79.2s\\
		2000&451.87&81.72s &\textbf{358.39}&84.38s  &412.55 & 100.7s  & 455.54 &121.62s\\
		5000&1138.98&188.78s&\textbf{895.96}&168.11s&1033.67& 234.43s & 1148.52 & 300.47s\\
		\bottomrule
	\end{tabular}}} 
\end{table}

\begin{table}[!t]
	\caption{Evaluation on Sliding-window Sampling (Dataset II) } 
	\label{tab:hyperparameter_2}\centering
		{\small
		\resizebox{10.5cm}{!}{
	\begin{tabular}{|c|cc|cc|cc|cc|}
		\toprule
		\multirow{2}*{Job No.}  & \multicolumn{2}{|c|}{$\beta=25$} &  \multicolumn{2}{|c|}{$\beta=50$} & \multicolumn{2}{|c|}{$\beta=100$} & \multicolumn{2}{|c|}{$\beta=250$}\\
	~ & Makespan & Time & Makespan & Time & Makespan & Time & Makespan & Time \\
		\midrule
		100& 55.33 &37.31s  & 54.11 & 32.94s & \textbf{37.45} & 27.89s & / &/\\
		250&125.79 &51.65s & 118.43&51.6s &\textbf{97.63} & 37.42s & 106.14 & 55.59s \\
		500& 235.34 &70.82s & 226.79 &68.15& \textbf{186.32} &42.81s & 211.73 & 91.17s \\
		1000& 445.67 &97.72s&  430.24 & 102.55s & \textbf{399.28} & 62.52s & 415.33 & 125.93s \\
		2000& 945.22 & 134.92s & 875.23 & 133.9s & \textbf{790.95} & 96.43s & 846.27 & 255.4s  \\
		5000& 2474.61 & 331.72s & 2467.37 & 314.33s & \textbf{1930.01} &171.79s & 2079.45 & 511.38s\\
		\bottomrule
	\end{tabular}}} 
\end{table}

From the above results, we can conclude that sliding-window sampling can help to improve the solution quality as well as the computational efficiency. 

%===============
\section{Conclusions}~\label{sec:con}

In this paper, we propose a bilevel constrained deep reinforcement learning model, \textit{BDS}, to solve the industrial scheduling problem in the practical automatic assembling warehouses. We make use of Double Deep Q Network (DDQN) as the upper level for efficient global sequence initialization and graph pointer network (GPN) as the lower level for high quality solution refining. We also design a sliding-window sampling mechanism to connect these two levels. Theoretical proof has been given to guarantee the algorithmic convergence. Experiments show that BDS significantly outperforms the other baseline methods. It should be emphasized that BDS is not limited to the industrial scheduling problem described in this paper, this bilevel structure can be used for many other large scale job shop scheduling or flow shop scheduling problems as well. In the future, we plan to fit the bilevel structure with some other deep reinforcement learning models besides DDQN and GPN.


%% New version of the num-names style

\bibliographystyle{main}
\bibliography{main.bib}
% \addbibresource{main.bib}






% \end{thebibliography}

\end{document}